# Advanced Predictive Quality Assessment for Ultrasonic Additive Manufacturing with Deep Learning Model

Lokendra Poudel*[1], Sushant Jha[2], Ryan Meeker[2], Duy-Nhat Phan[1], Rahul Bhowmik*[1]

[1]Polaron Analytics, 9059 Springboro Pike, Miamisburg, OH, 45342, USA

[2]University of Dayton Research Institute, 300 College Park, Dayton, OH, 45469, USA

*Correspondence to:

Email address: lokendra@polaronanalytics.com
              rahulbhowmik@polaronanalytics.com

**Abstract**

Ultrasonic Additive Manufacturing (UAM) employs ultrasonic welding to bond similar or dissimilar metal foils to a substrate, resulting in solid, consolidated metal components. However, certain processing conditions can lead to inter-layer defects, affecting the final product's quality. This study develops a method to monitor in-process quality using deep learning-based convolutional neural networks (CNNs). The CNN models were evaluated on their ability to classify samples with and without embedded thermocouples across five power levels (300W, 600W, 900W, 1200W, 1500W) using thermal images with supervised labeling. Four distinct CNN classification models were created for different scenarios including without (baseline) and with thermocouples, only without thermocouples across power levels, only with thermocouples across power levels, and combined without and with thermocouples across power levels. The models achieved 98.29% accuracy on combined baseline and thermocouple images, 97.10% for baseline images across power levels, 97.43% for thermocouple images, and 97.27% for both types across power levels. The high accuracy, above 97%, demonstrates the system's effectiveness in identifying and classifying conditions within the UAM process, providing a reliable tool for quality assurance and process control in manufacturing environments.

## 1.0 Introduction

Additive manufacturing (AM) refers to a set of computer-controlled techniques that create three-dimensional objects by layering materials (Ansari et al., 2022; Saimon et al., 2024). Ultrasonic additive manufacturing (UAM) is a standout solid-state manufacturing method within this group, producing nearly finished metal parts without melting the materials. It uses ultrasonic energy and pressure to create a bond between layers, enabling the creation of complex parts and even embedding electronics like sensors during the printing process (Behvar et al., 2024; Bournias-Varotsis et al., 2019; Han et al., 2022; Hehr & Norfolk, 2019). This capability sets UAM apart from other methods like laser powder or wire-feed techniques, which can't embed sensors during printing. One of the major advantages of UAM is its ability to print different metals together and incorporate all necessary components and sensors into a single module (Bournias-Varotsis et al., 2019). This results in greater design flexibility and less material waste compared to traditional manufacturing. UAM's potential is being explored in various fields, including aerospace, satellite technology, automotive, and biomedical engineering (Behvar et al., 2024). Another key benefit of UAM is its suitability for printing in space, as it doesn't involve melting materials or dealing with the challenges of a melt pool, making it ideal for on-orbit manufacturing (Hehr & Norfolk, 2019; Reitz et al., 2021). Research in in-situ monitoring and control in UAM is expanding quickly because it enables the production of high-quality, defect-free parts. Despite the attractive features and potential for industrial adoption, industries still face challenges in using additive UAM as their primary method for mass production. The main issues are inconsistencies in part quality and properties, along with process inefficiencies (Behvar et al., 2024; Nadimpalli et al., 2020).

The quality of parts produced by UAM depends on several factors, including the vibration amplitude, clamping force, welding speed, substrate temperature, and the surface conditions of the materials and tools used (Nadimpalli et al., 2020). Quality control of mechanical components is crucial to ensure their expected performance and prevent their failure. For components manufactured additively, quality control performed in-process is particularly interesting, as the sequential deposition and rebuilding of layers represent a possibility to mitigate existing flaws. The first step towards closed-loop control is to ensure that the monitoring setup and the data analytics approach can flag and discriminate flaws (Schwerz & Nyborg, 2022; Zhang & Zhao, 2022). However, analyzing the vast amounts of data generated during the process in real-time is a significant challenge with traditional statistical methods (Saimon et al., 2024). Deep learning (DL), on the other hand, excels at quickly identifying patterns and extracting important information from this data. This capability makes it an effective tool for implementing real-time quality control in UAM processes. Convolutional neural networks (CNNs), a type of DL, combine advanced image processing with deep learning capabilities. They are particularly powerful due to their efficient automatic feature extraction, outperforming traditional machine learning (ML) models like decision trees, random forests, and support vector machines (SVMs) in both accuracy and training time (Ansari et al., 2022). They are highly scalable and adept at handling large datasets. Their scalability and efficiency further enhance their suitability for large image-based analysis tasks, reinforcing their role as a superior choice in ML for monitoring quality control process of UAM.

Image-based monitoring is one of the most widely used methods in AM due to the ability of DL algorithms, like CNNs, to extract spatial features from grid-like data and the availability of high-resolution cameras (Kim et al., 2023; Ördek et al., 2024; Song et al., 2020). Studies applying DL methods to image-based monitoring can be grouped into two main categories: DL-based image processing and deficiency monitoring for quality assessment. DL-based image processing involves using DL techniques to manipulate, transform, or extract features from input images to improve UAM process monitoring. High-quality input data is crucial for achieving optimal monitoring performance. Techniques such as image augmentation, segmentation, transfer learning, and feature extraction are commonly employed (Ansari et al., 2022; Xia et al., 2022). Deficiency monitoring uses for quality assessment focuses on detecting and evaluating imperfections like defects, anomalies, deformations, faults, or cracks during the UAM process to ensure the quality of the printed product (Tian et al., 2020; Yangue et al., 2023). However, DL-based image monitoring needs large volumes of high quality, well-labelled data and significant computational resources, which can be costly and energy intensive. DL models are often complex and lack interpretability, making real-time processing and integration with UAM systems difficult. They may struggle to generalize across different processes and conditions, and their performance can degrade without continuous learning and updates.

In practice, the layer-wise printing process can be influenced by variations due to the highly nonstationary nature of the process dynamics, leading to weld anomalies in some locations. Furthermore, surface defects that occur in one layer are likely to propagate to subsequent layers, potentially on a larger scale, which can severely compromise the overall quality of the printed product. Therefore, it is essential to have an efficient and effective solution for layer-wise surface quality assurance in UAM. This necessity drives the development of a data-driven model for predicting layer-wise surface morphology in printed parts. This study aims to design an effective layer-wise monitoring system combined with a supervised DL approach to identify and classify defects in UAM processes under different welding powers. To the best of our knowledge, this is the first study to propose employing deep learning in the UAM process. The method relies on thermal images collected from in-situ monitoring of specimens produced with different power settings, thereby containing varying defect levels. Since defects are systematically generated, they are distributed throughout the material with systematic variation in occurrence rate. Each monitored layer is used to train a neural network, which then classifies images to identify primary features including physical weld defects and anomalous regions due to power variations during printing. This system's ability to detect multiple flaw types enhances control efficiency, highlighting the importance of assessing its capability to identify different flaws.

The remaining sections of the paper are structured as follows: Section 2, Research Methodology, describes the experimental setup for data generation and analysis, and the deep learning techniques used for defect classification with neural network structures. Section 3, Results and Discussions, presents the findings and their analysis. Section 4, Conclusions, summarizes the study and proposes future research directions.

## 2.0 Research Methodology

### 2.1 Experimental setup for data generation

The chosen material for the part was the aluminum alloy Al 6061-T651, selected for its relevance to aerospace and manufacturing applications (Waller et al., 2019). In addition to its low cost, Al 6061 has high strength to weight ratio, improved corrosion resistance, machinability, and weldability properties (Waller et al., 2019), which makes it attractive for aerospace structural applications. Due to its importance in aerospace, Al 6061 is a candidate material for additive manufacturing (Roberts et al., 2016) including UAM (Sridharan et al., 2016).

To advance our DL-based defect detection capability within the UAM process, we executed a comprehensive data collection endeavor utilizing Fabrisonic SonicLayer 1600 UAM machine equipped with 35 KHz horn or welding head (**Fig. 1**). The horn is a crucial component of the UAM machine (see **Fig. 1 (a), (b)**) that applies ultrasonic vibration and pressure required for solid state welding between material layers. The quality of the printed coupons is controlled through various horn parameters, including oscillation amplitude, normal force, travel speed, and surface condition. Our focus was on printing components under realistic conditions along with reasonable deviations from optimal conditions, allowing us to gather in-situ sensor datasets essential for this study. To investigate the defect characteristics induced by power variation of welding with/without embedded electronics, we strategically embedded a thermocouple within the part during printing. Additionally, we installed a near-infrared thermal imaging camera inside the UAM machine to capture thermal images throughout the build process (see **Fig. 1(c)**). The thermal imaging camera chosen for this study was an Optris PI 640i G7 infrared camera. Its specifications include a low energy range of 0 – 250°C and an accuracy of ±2°C or about ±2%. The camera captures 16-bit images at 32 frames per second and a resolution of 640x480 pixels. In addition, the camera was equipped with the O33 standard lens, which was focused appropriately to the weld head to capture images under optimal configuration. This comprehensive approach enabled us to collect invaluable data, essential for analyzing defect patterns, thermal dynamics, and refining our machine learning algorithms for enhanced defect detection accuracy.

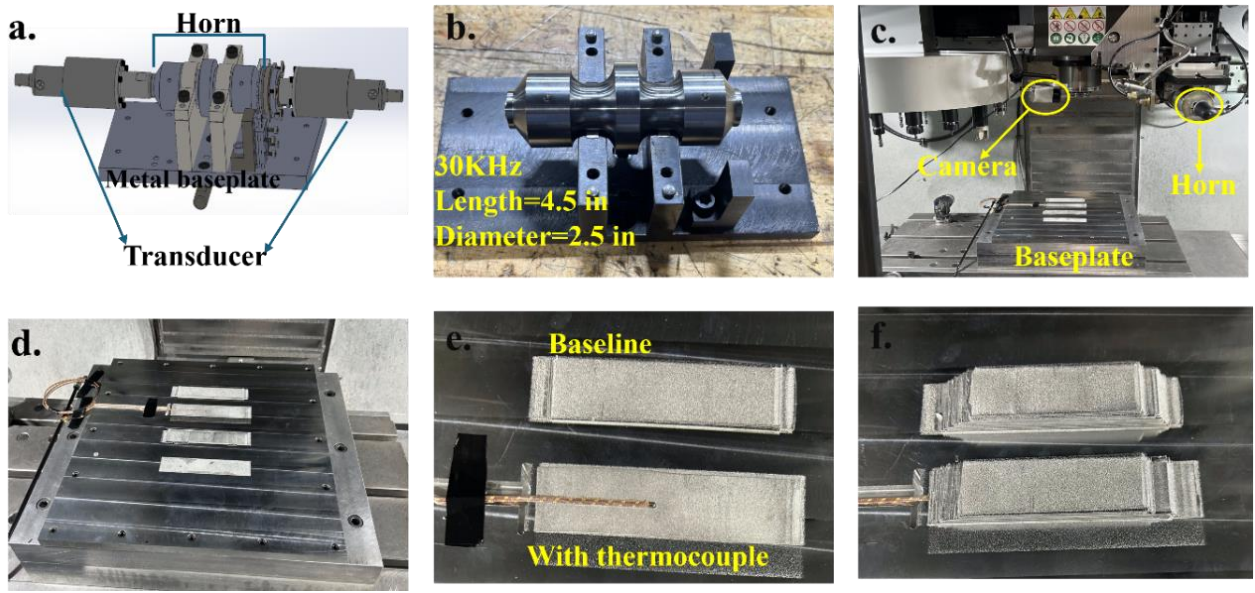

Figure 1: Experimental data collection via UAM process at Fabrisonic LLC: (a) A CAD design of Horn, (b) 35 KHz horn, (c) SonicLayer 1600 Machine, (d) a close-up of the build plate (f) the generation of layers on the build plate, and (g) the final printed coupons

For generating thermal images, we printed two coupons: one without a thermocouple (baseline) and one with a thermocouple (see **Fig. 1(e), (f)**). During printing process of both coupons, we varied the welding power every 10 layers, starting at the regular power level of 900 W. We then applied lower power levels of 600 W and 300 W, followed by higher power levels of 1200 W and 1500 W.

**2.2 Data processing and Analysis**

Applying image processing is optional when training a CNN model, but it can be beneficial for reducing data complexity, eliminating noise or background, and enhancing visual appeal. We utilized several current algorithms for data processing, including vertical and horizontal flipping, cropping in height, width, and time dimensions, normalization, and principal component analysis (PCA) (see **Fig. 2(a)** for workflow of data processing).

During our CNN training, we cropped the image to focus on the welding head and removed frames before and after the welding occurred. Normalization adjusts the pixel values in the video to a specified range, making it useful for algorithms sensitive to pixel value ranges or for visualization frameworks that expect a particular pixel range. PCA serves as both an image enhancement and compression tool. It simplifies noise reduction in the image, helping CNN focus on more relevant features. PCA reshapes the video into a two-dimensional matrix by converting each frame into a column, then generates the covariance matrix and factorizes it using singular value decomposition (SVD). This process separates the video into components ordered by variance. High-variance components contain more information, while low-variance components are mostly noise. By eliminating some low-variance components and reversing the PCA process, we can de-noise and reconstruct the video. The parameter for the number of components to retain defaults to 0.8, meaning 80% of the components are kept, reducing noise by 20%. Additionally, this denoising makes the video easier to compress, significantly shrinking its size when encoded in a standard format like mp4.

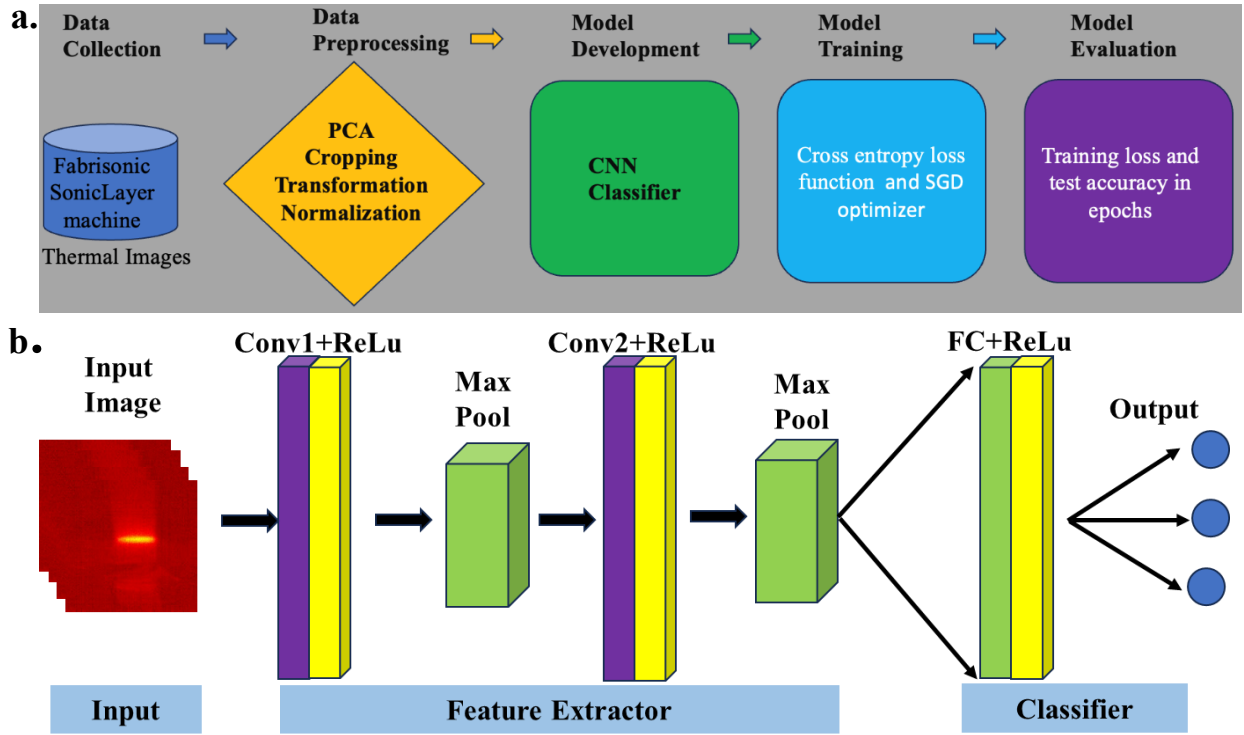

Figure 2: (a) The workflow of CNN model development for thermal image analysis, and (b) Architecture of CNN classifier.

## 2.3 Deep learning algorithm for automatic defect detection

To tackle defect detection and classification, a supervised DL approach is utilized. DL has gained popularity in machine vision and pattern recognition because of its excellent capabilities in feature learning and image classification (Xia et al., 2022). Among the various DL algorithms, convolutional neural networks (CNNs) stand out as one of the most widely used. CNNs have been successfully implemented in a variety of applications, including object detection, action recognition, and image classification. This study involves creating an annotated image dataset for training, validating, and testing a trained CNN. A typical CNN has two main components: a feature extraction network and a fully connected network (FCN) (see **Fig. 2(b)**). The feature extraction network combines convolutional and pooling layers with non-linear activation functions. For the CNN classifier, we used convolutional layers with various kernel sizes that move across the input features, extracting a range of features through convolution. The fully connected network (FCN) processes these features through hidden layers of neurons, which are learned from the extracted features (see **Table 1**) for the CNN classifier parameters). The final and last hidden layers of the FCN are linked to a ReLU regression for classification. During training, the network compares its output with the desired results, computes gradients, and backpropagates these gradients to adjust the weights in the convolutional layers. This iterative process continues with new data batches until the output error falls below a certain threshold.

Table 1: Layer and parameter of CNN classifier models

| Layer No. | Layer-Type | Parameters | | |
|---|---|---|---|---|
| 1 | Convd2d | Kernel size | Stride | Dilation |
| | | 3 | 1 | 1 |
| 2 | Activation | ReLU | | |
| 3 | Maxpool2d | Kernel size | Stride | Dilation |
| | | 2 | 2 | 1 |
| 4 | Cond2d | Kernel size | Stride | Dilation |
| | | 3 | 1 | 1 |
| 5 | Activation | ReLU | | |
| 6 | Maxpool2d | Kernel size | Stride | Dilation |
| | | 2 | 2 | 1 |
| 7 | Flatten | Linear | | |
| 8 | Activation | ReLU | | |
| 9 | Output | Linear | | |

### 3.0 Results and Discussion

### 3.1 *Ex-situ* microstructure characterization

In this study, we compared the quality of welding achieved through high to low-power UAM by examining the microstructure of the printed coupons. High-power UAM, typically involving energy levels of 1000W and above, provides greater energy and amplitude in ultrasonic vibrations (Hehr & Norfolk, 2019). On the other hand, low-power UAM, generally using energy levels below 900W, applies less energy and amplitude. Two coupons were printed at Fabrisonic using varying power levels, with one including a thermocouple and the other without the thermocouple. In both coupons, the welding power was adjusted every 10 layers, starting at the standard power level of 900 W. Subsequently, lower power levels of 600 W and 300 W were used, followed by higher power levels of 1200 W and 1500 W (see Figure 3(d) for the layer and power level diagram). These coupons were then subjected to microstructural analysis to evaluate the quality of the welding. For this analysis, the samples were sectioned along the center, and the plane containing the build direction (i.e., along the height) and width of the coupon was characterized (see Figures 3(a), (b), and (c)). The extracted sections were mounted and metallographically prepared using grinding and fine polishing procedures. The metallographic samples were then characterized using an optical microscope. The optical images revealing the void distribution in the baseline coupon and the coupon with embedded thermocouple are shown in Figures 3(e) and (f), respectively. The voids appear to be weld defects oriented along the interface between layers attributed to insufficient energy input at those locations (Sridharan et al., 2016).

In the optical microscopy images shown in Figures 3(e) and (f), the power setting varies along bottom to top direction. The microstructural analysis shows that both samples contain a higher volume fraction of voids or pores when lower power levels (300W and 600W) are used, whereas fewer voids are observed at higher power levels (1200W, and 1500W). At standard power level (900W), voids volume fraction is higher than under power levels of 1200W and higher. The voids fraction is highest at lower power levels, 600W and 300W. Therefore, the standard power of 900W seems to be a tradeoff between power level and voids volume fraction. These results indicate that power level is crucial for the quality of printed samples, as higher power enhances weld consolidation and interface strength. However, very high power may cause excessive deformation and debonding (Sridharan et al., 2016). Power failures or fluctuations can significantly affect the quality of parts, highlighting the importance of automatic detection of power changes and defects in manufacturing facilities. Thus, integrating computational methods for automatic identification of regions in the part influenced by power failures or fluctuations and defects induced by power variations is essential to prevent manufacturing issues. To address this, we developed DL-based classification models built on the applied powers during the printing of coupons. Further details are provided in the following sections.

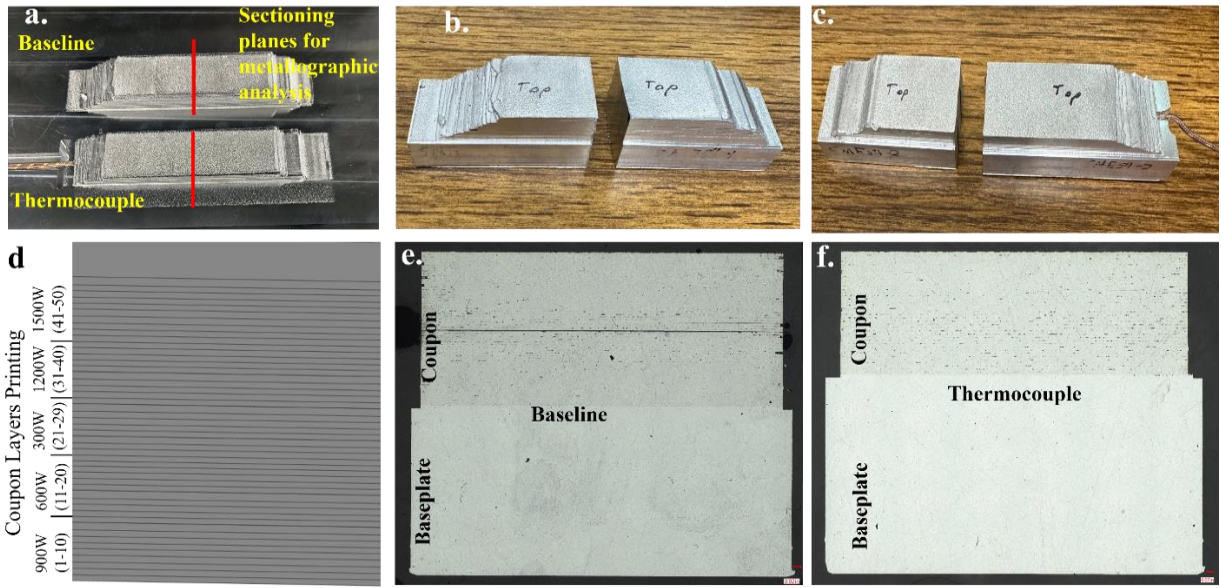

Figure 3: Metallographic analysis of two specimens (baseline and thermocouple) is presented. (a) Illustration of sectional planes for analysis, (b) baseline coupon, (c) thermocouple coupon, (d) Sketch of x-y plane of coupon layers printing in applied power levels, microstructure image showing voids in baseline sample, and (e) thermocouple sample.

## 3.2 Construction of classification datasets

The datasets used to train, validate, and test the CNN were created using thermal images captured layer-by-layer from the two specimens (baseline and thermocouple) with layer dimensions of approximately 6 inch long and 1 inch wide. The parameters employed were layer thickness of 0.006 inches, welding speed of 80 inches per minute, and power levels of 300W, 600W, 900W, 1200W, and 1500W. Thermal images with dimensions of 640×480 pixels were captured from the specimens during layer wise printing and were labeled according to the predefined classes using power settings for UAM processing. To generate the datasets, we focused on collecting images specifically during the welding period and removed any frames captured before and after the welding occurred. A total of 2760 labelled images were obtained, out of which 1389 were assigned the label "baseline" and 1371, labeled as 'thermocouple". Prior to CNN training, we cropped the images (160×160 pixels) to focus on the welding head. This was followed by preprocessing steps such as normalization and PCA. Sample images obtained from two specimens of raw and preprocessed images are presented in **Figure 4**. As shown, we categorized the collected images based on the five power levels.

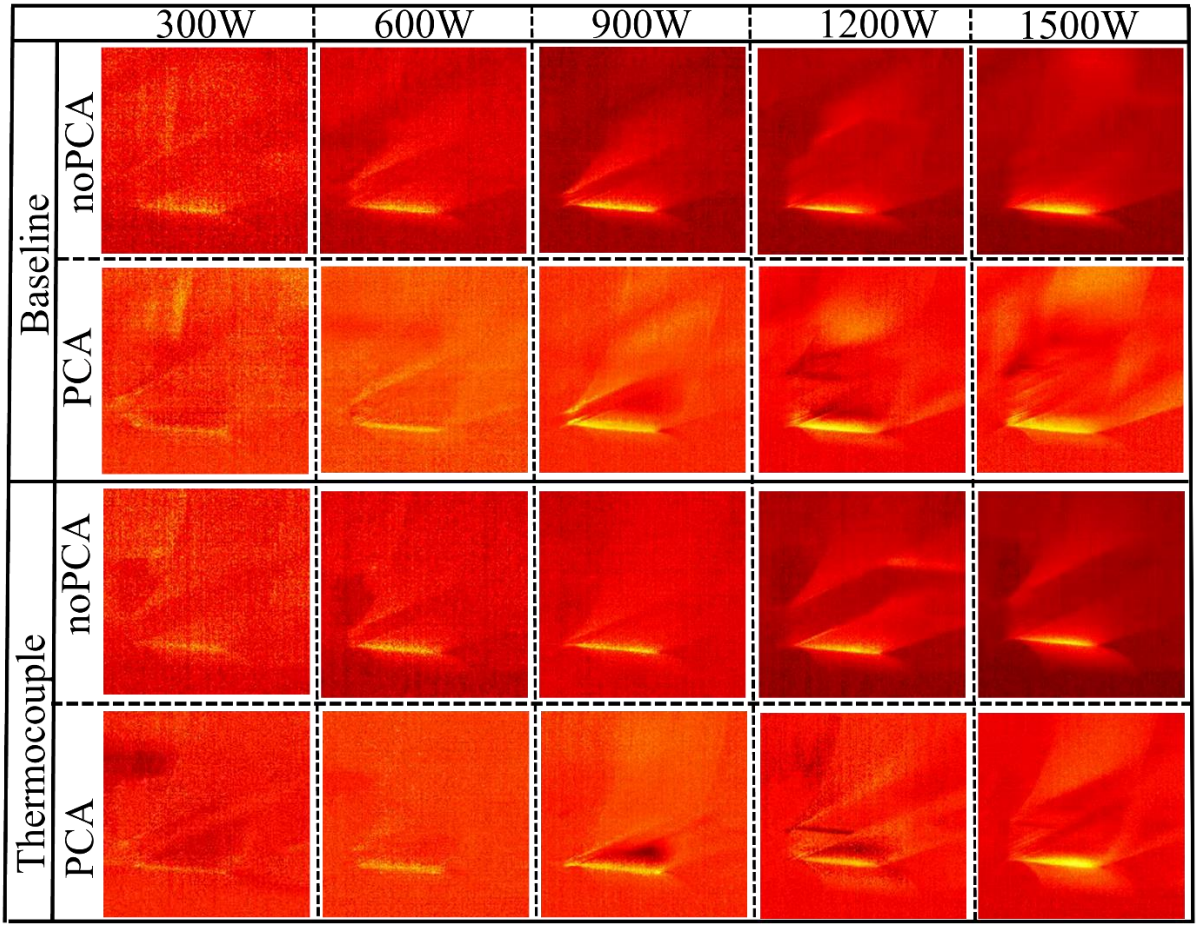

Figure 4: Illustrated sample thermal images of two specimens (baseline and thermocouple) with distinct power levels (300W, 600W, 900W, 1200W, and 1500W), shown both after cropping and with/without PCA applied.

### 3.3 CNN Model development

A CNN model was designed to classify thermal images per primary classes assigned through coupon printing. This choice was made because CNN can process inputs of any dimensions, even though our training is conducted on images of a fixed size. This architecture is suitable for semantic segmentation, allowing the trained network to be applied to any part geometry. Training and validation were performed on the dataset constructed as described in the previous section. The network layers were designed to handle the 160×160 inputs and reduce the dimensions gradually through convolution and max-pooling operations, followed by a final fully connected layer that results in an output of dimensions 1×1×n (n: number of classes). Therefore, the fully connected layer typically presents in the last steps of a neural network that produces an output of dimensions that are dependent on the number of classes. This layer is then used as an input to the ReLU layer. Finally, a classification layer is used to output a predicted class. The network was designed to contain adequate complexity to avoid overfitting while being relatively simple. For that, a limited number of parameters were used, still ensuring satisfactory validation accuracy. Moreover, two max-pooling layers were used to decrease spatial resolution, thereby decreasing computational time. The network is schematically visualized in **Figure 2(b),** and its parameters, including kernel size, stride, padding, and dilation, are detailed in **Figure 2(c).** In this study, we developed four distinct CNN models, each tailored to a specific task. Model_1 was designed to classify baseline and thermocouple prints using a standard power level of 900 W. Model_2 focused on classifying baseline prints across five different power levels: 300 W, 600 W, 900 W, 1200 W, and 1500 W. Model_3 was similarly tasked with classifying thermocouple prints at these five power levels. Finally, Model_4 aimed to classify both baseline and thermocouple prints across the five power levels, resulting in a total of 10 classes.

## 3.4 Model training and performance evaluations

The quality of parts is influenced by the welding power during printing. The primary objective of CNN model is to classify data into five distinct power level printing scenarios for baseline and with embedded thermocouple printing. For model training, the dataset was split into training and test data at the ratio of 80% and 20%, respectively. Class imbalance is one of the most common challenges in data-driven solutions. It occurs when there is a significant difference in the number of instances of one class compared to others within the dataset. In such cases, using accuracy as an evaluation metric can be misleading if the data are biased towards a particular class, leading to a biased model. To address this, we ensured that the training datasets for each class were balanced in this study. The training data distribution is presented in **Figure 5.**

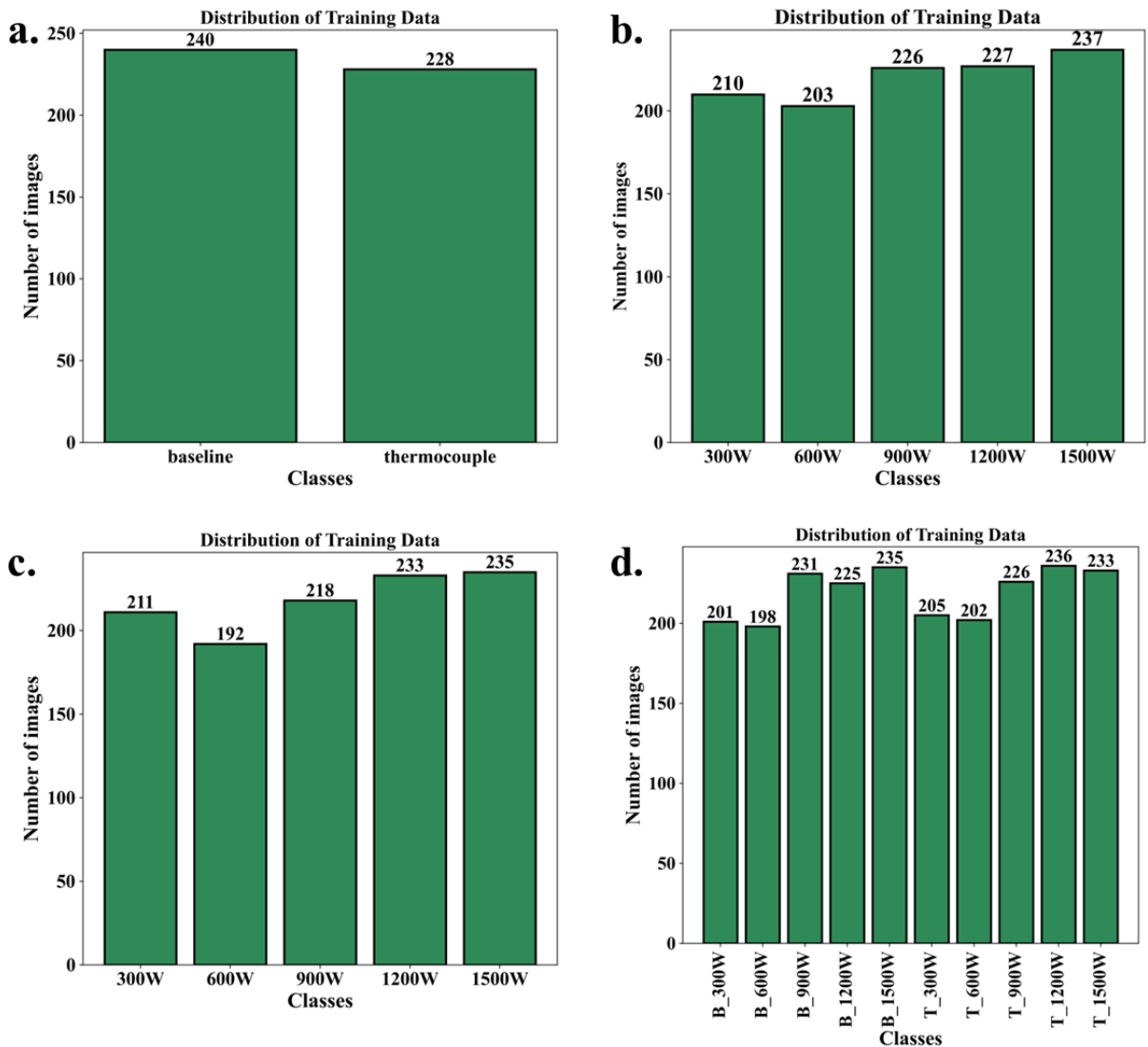

Figure 5: Training data distribution for developed CNN models in (a) Model_1, (b) Model_2, (c) Model_3, and Model_4.

The model training was performed with cross-entropy loss function, stochastic gradient descent (SGD) with fixed momentum of 0.9 and learning rate of 0.001 using batch size 16 up to 50 epoch. The training loss and test accuracies of the developed models are presented in **Table 1**, while the plots showing the models' training progress are depicted in **Figure 6**. The model achieved an accuracy of 98.29% on baseline and thermocouple images combined. For baseline images classified into five different power levels, the accuracy was 97.10%. For thermocouple images with the same power level classification, the accuracy was 97.43%. When both baseline and thermocouple images were considered across the five power levels, the accuracy was 97.27%. These results demonstrate that the developed CNN models provide highly accurate predictions for UAM printing layers.

Table 1: CNN models and their test accuracy and training loss on the best parameter's values

| Model | Test Accuracy | Training Loss | #parameters | Hyperparameters |
|---|---|---|---|---|
| Model_1 | 98.29 | 0.0903 | 13,126,978 | lr=0.01, momentum=0.9, epoch=50 |
| Model_2 | 97.10 | 0.0225 | 13,127,235 | lr=0.01, momentum=0.9, epoch=50 |
| Model_3 | 97.43 | 0.0115 | 13,127,365 | lr=0.01, momentum=0.9, epoch=50 |
| Model_4 | 97.26 | 0.0061 | 13,128,010 | lr=0.01, momentum=0.9, epoch=50 |

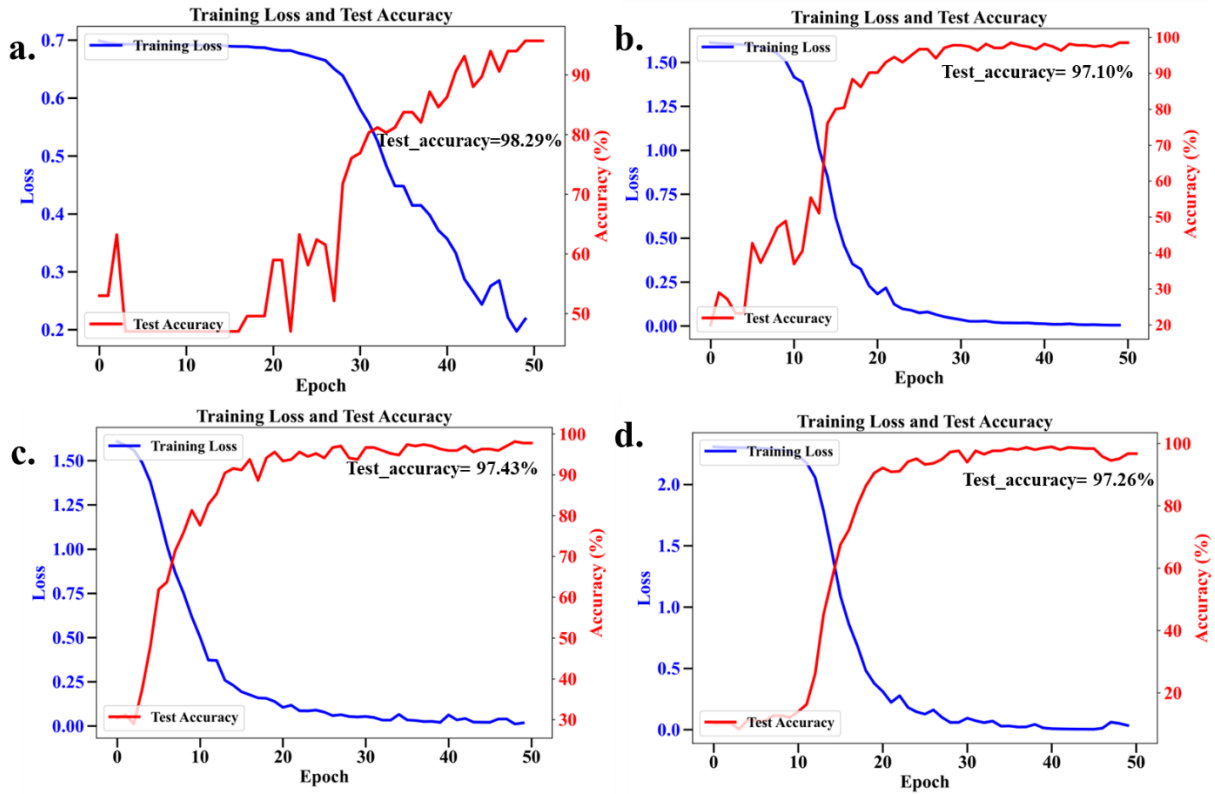

Figure 6: Training loss and testing accuracy progress plots for developed CNN models in (a) Model_1, (b) Model_2, (c) Model_3, and Model_4.

The model was evaluated using an accuracy metric based on confusion matrix. Confusion matrix is a technique for evaluating the performance of a model by analyzing the model's categorical prediction for false positives (FP), false negatives (FN), true positives (TP), and true negatives (TN). Considering a binary problem (positive and negative classes), TP and TN are correct positive and negative classifications, respectively. FP is an incorrect positive prediction when it is negative, and FN is an incorrect negative prediction when it is positive. Therefore, a FN is more dangerous than a FP in defect detection. In the confusion matrix representation, the model's accuracy is represented by the elements on the main diagonal, corresponding to TP and TN. Therefore, it provides a clear visual representation of the model's successes and errors, highlighting correct predictions and discrepancies between the model's predictions and the true (actual) classes of the data. In this work, for the class detector approach, classes are represented by "baseline" and "with thermocouple", respectively. Moreover, accuracy is a statistical measure that allows the evaluation of the performance of deep learning models in classification tasks. It is the proportion of true positives and true negatives over the total number of predictions made. Therefore, this metric indicates the model's ability to correctly assign labels to object categories. Mathematically, accuracy (A) is defined as:

$$Accuracy(A) = \frac{TP+TN}{TP+TN+FP+FN} \quad \ldots\ldots\ldots\ldots..(1)$$

The Accuracy metric results in a value between 0 and 1, where values closer to 1 indicate better classification.

For evaluating the developed CNN models, we calculated and plotted the confusion matrix to assess their classification performance. The initial information can be obtained through a binary classification problem, where each object is categorized into one of two classes: positive (baseline) or negative (thermocouple) (see **Figure 7(a)**). The second type of information, regarding power levels, can be addressed using a multiclass classification method, where each object is assigned to one of five classes, each representing a different power level used for printing the sample layers (see **Figure 7(b) & (c)**). The third type of information, which includes both sample types and power levels, can also be treated as a multiclass problem. Here, data is classified into ten classes, with each class representing a specific power level for both baseline and thermocouple coupons (see **Figure 7(d)**). The confusion matrix provides a more detailed explanation of the predictions of the developed CNN models for thermal images in UAM process. The confusion matrix results showed that the performance of all model's performance in the classification tasks exceeded 97.00%.

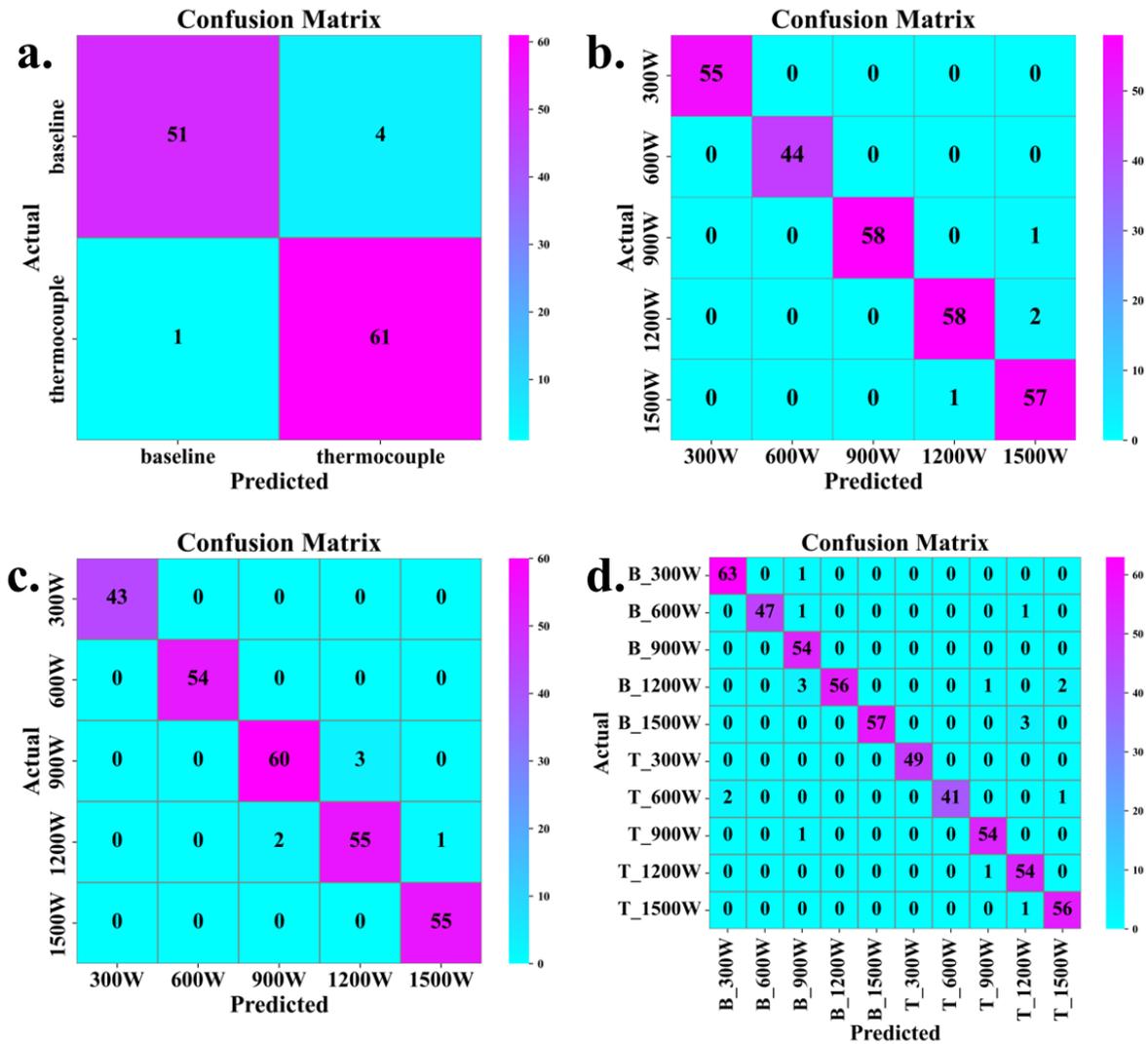

Figure 7: Confusion matrix showing the performance of the developed CNN models in classifying images, (a) Model_1, (b) Model_2, (c) Model_3, and Model_4.

## 4.0 Conclusions and future perspective

This paper introduces a novel approach to applying a CNN model for manufacturability analysis. The CNN model was trained and shown to be capable of identifying the predominant printing class type based on thermal images obtained through in-situ monitoring. Machine learning techniques were employed to ensure the model's robustness. The study demonstrated that the CNN, trained on labeled thermal images, can accurately classify specimen layers based on whether they were printed without a thermocouple (baseline) or with a thermocouple, and into various power levels categorized as 300W, 600W, 900W, 1200W, and 1500W. The model achieved a high accuracy of 97.26% on the test set for classifying specimen layers according to these criteria, with all models consistently achieving testing accuracy above 97% when systematically varying UAM printing power levels were applied.

There are also some limitations to this study. First, the dataset is still limited, which is a common issue in most machine learning applications within UAM. To train a machine learning model with reliable results, it is generally recommended to have a large dataset. A small dataset can lead to overfitting, meaning the proposed model might not generalize well to the broader population. To address limited datasets, we can apply techniques like data augmentation, transfer learning, and synthetic data generation to enhance data diversity and model performance. Additionally, methods such as cross-validation, regularization, and few-shot learning help improve model generalization and reduce overfitting. Currently, available UAM datasets are very limited, and the UAM process itself is not fully standardized. There is a research gap in establishing a specific data framework for UAM, including what information should be stored in the dataset and how it should be standardized. Secondly, generating thermal images with all the necessary parameters and features is a time-consuming process. To achieve a more accurate understanding of the UAM process, it is crucial to create more comprehensive datasets using various parameters, such as ultrasonic oscillations, strain rate, horn speed, metal types or various alloys, and printing coupons at different power levels separately. In the future, we plan to conduct further studies on weld development in vacuum (on-orbit) environments, both with and without electronics embedding, exploring various power levels and different metal types or their combinations. We will apply advanced machine learning techniques such as transfer learning, synthetic data generation, cross-validation, and few-shot learning to enhance our analysis.

One of the main benefit of the CNN approach is its ability to leverage transfer learning, allowing a pre-trained model to be adapted to address new, similar challenges with minimal re-training (Vesala et al., 2022; Zhu et al., 2023). Regarding potential industrial applications, the research conducted here is valuable for various sectors, including aerospace, automotive, on-orbit and biomedical engineering. UAM offers advantages such as increased design flexibility and reduced material waste, making it a promising method for manufacturing complex metal parts. By examining the defects and microstructure of UAM-processed Al6061-T651 foils, this study contributes to the exploration of potential uses of DL with UAM in these industries. It demonstrates the UAM system's effectiveness in identifying and classifying conditions within the UAM process, providing a reliable tool for quality assurance and process control automatically in manufacturing environments. Finally, this study determined a significant correlation between the electronics embedded in the printing process, the power of welding, and the resulting microstructural changes in the UAM process. This understanding of material behavior and performance can inform the optimization of UAM parameters and process control, leading to enhanced quality, reliability, and efficiency in the manufacturing of metal parts.

**Declaration of Competing Interest**

The authors declare that they have no competing financial interests or personal connections that might have influenced the results of this research.

**Data availability**

Experimental data and developed codes will be made available on request.

**Ackowledgements**

This material is based upon work supported by the Air Force Research Laboratory, AFWERX, AFRL/RGKB under Contract No. FA864923P1242. Any opinions, findings and conclusions or recommendations expressed in this material are those of the authors and do not necessarily reflect the views of the Air Force Research Laboratory, AFWERX, AFRL/RGKB. The manuscript was cleared for public release on 10/04/2024, with Case Number: AFRL-2024-5576. The authors would also like to acknowledge Mark Norfolk, Jason A. Riley, and Matthew Burkhart of Fabrisonic LLC.